\documentclass{WileyMSP-template}
\usepackage{url}
\usepackage{float}
\usepackage{hyperref}
\usepackage{xcolor}
\usepackage{comment}
\begin{document}

\pagestyle{fancy}
\rhead{\includegraphics[width=2.5cm]{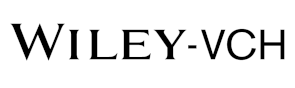}}

\title{A Perspective on Explainable Artificial Intelligence Methods: SHAP and LIME}

\maketitle


\author{Ahmed M Salih*,}
\author{Zahra Raisi-Estabragh,}
\author{Ilaria Boscolo Galazzo,}
\author{Petia Radeva,}
\author{Steffen E. Petersen,}
\author{Karim Lekadir$^\ddagger$,}
\author{Gloria Menegaz$^\ddagger$}



\begin{affiliations}

Dr. A. M. S., Dr. Z. R., Prof. S. E. P.\\
William Harvey Research Institute, NIHR Barts Biomedical Research Centre, Queen Mary University of London, London, UK\\
Barts Heart Centre, St Bartholomew’s Hospital, Barts Health NHS Trust, West Smithfield, London, UK \hfill \break

Dr. A. M. S.\\
Department of Population Health Sciences, University of Leicester, Leicester UK\\
Department of Computer Science, Faculty of Science, University of Zakho, Zakho, Kurdistan Region, Iraq\\
Email Address: a.salih@leicester.ac.uk \hfill \break

Dr. I. B. G, Prof. G. M.\\
Department of Engineering for Innovation Medicine, University of Verona, Verona, Italy\hfill \break

Prof. P. R., Dr. K. L.\\
Department of de Matemàtiques i Informàtica, University of Barcelona, Barcelona, Spain\hfill \break

Prof. S. E. P.\\
Health Data Research UK\\
Alan Turing Institute, London, UK\hfill \break

*: Correspondence author: a.salih@leicester.ac.uk.\\
$\ddagger$: These authors contributed equally to this work.
\end{affiliations}



\begin{abstract}
eXplainable artificial intelligence (XAI) methods have emerged to convert the black box of machine learning (ML) models into a more digestible form. These methods help to communicate how the model works with the aim of making ML models more transparent and increasing the trust of end-users into their output. SHapley Additive exPlanations (SHAP) and Local Interpretable Model Agnostic Explanation (LIME) are two widely used XAI methods, particularly with tabular data. In this perspective piece, we discuss the way the explainability metrics of these two methods are generated and propose a framework for interpretation of their outputs, highlighting their weaknesses and strengths. Specifically, we discuss their outcomes in terms of model-dependency and in the presence of collinearity among the features, relying on a case study from the biomedical domain (classification of individuals with or without myocardial infarction). The results indicate that SHAP and LIME are highly affected by the adopted ML model and feature collinearity, raising a note of caution on their usage and interpretation.

\end{abstract}

\section{Introduction}
Machine (ML) and Deep (DL) learning methods have shown great success in a variety of domains, e.g. biology~\cite{richards2022application}, medicine~\cite{hamet2017artificial}, economy~\cite{gyasi2019survey} and education~\cite{zhang2021ai}. However, such success is accompanied by complexity in understanding how these models work, why the models make a specific decision, what features/regions are most influencing the model output and the degree of certainty the model has in the generated outcome. All of these questions and more are raised by the end-users, specially when advanced models including deep neural networks are implemented. Accordingly, a new field of research has emerged named eXplainable artificial intelligence (XAI) aiming at demystifying 'black box' models into a more comprehensible form~\cite{gunning2019xai}. XAI is indispensable to increase the model transparency and the trust of end-users in the model outcome~\cite{szabo2022clinician}~\cite{ali2023explainable}. Such additional reassurances are essential for wide implementation of such models, particularly in high risk fields such as healthcare. However, specific aspects such as model-dependency and collinearity across the features might affect the quality of the XAI outcome. In this perspective paper, we aim at revealing how model-dependency and the presence of collinearity affect the XAI outcome. Moreover, we use a case study from the biomedical domain to examine the effects of the aforementioned issues on two of the most common XAI methods. In addition, another case study was used to reveal how the XAI methods can be implemented and what are the possible solutions to overcome their limitations in terms of model-dependency and collinear features.
\section{eXplainable artificial intelligence}
\begin{figure}[H]
\centering
\includegraphics[height=8 cm]{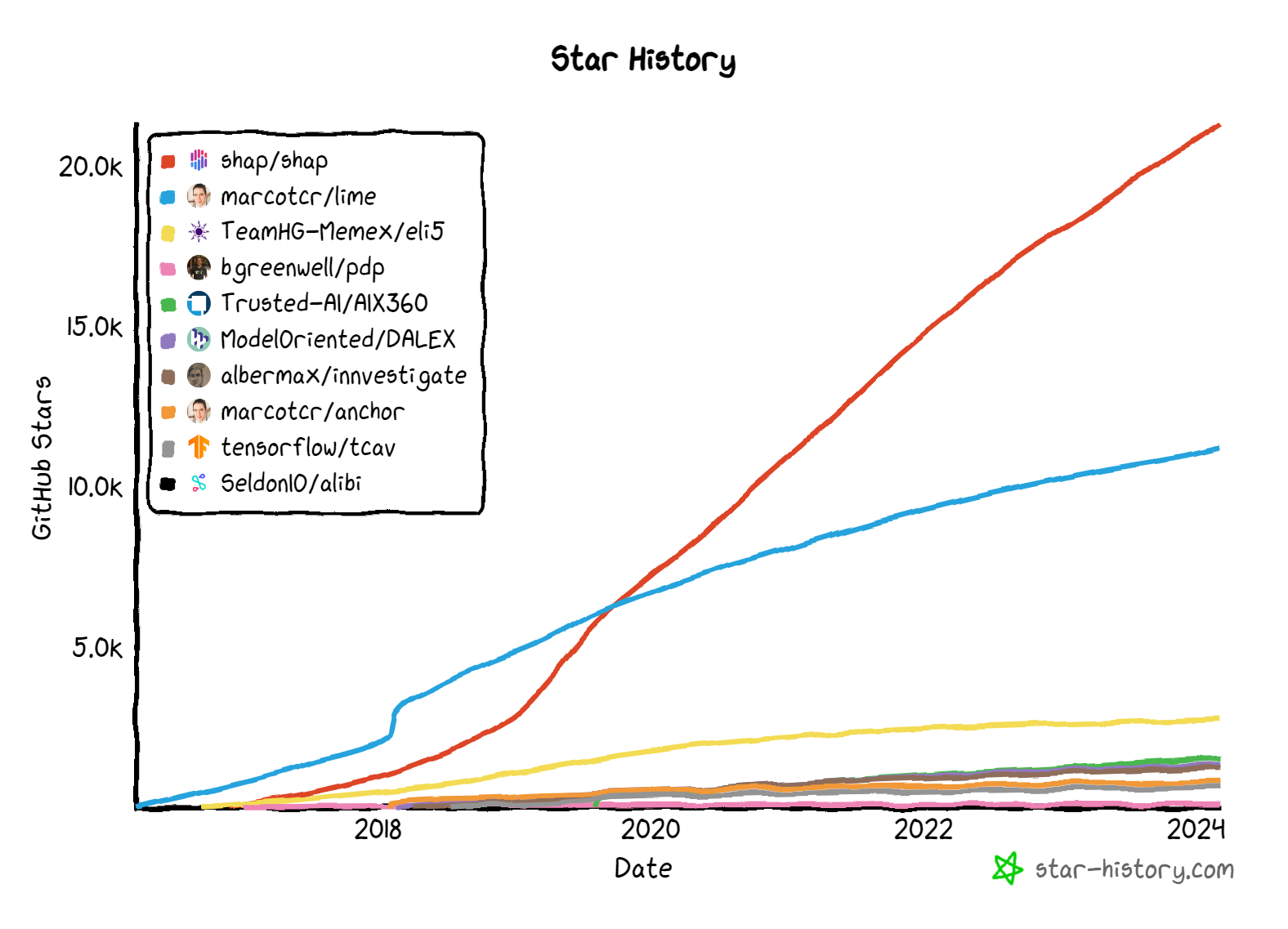}
\caption{GitHub Star for 10 common XAI methods.}\label{star}
\end{figure}

Several approaches have been proposed as XAI methods dealing with a variety of data and model types, aiming at explaining the models outputs locally and globally. Among these, SHapley Additive exPlanations (SHAP)~\cite{lundberg2017unified} and Local Interpretable Model Agnostic Explanation (LIME)~\cite{ribeiro2016should} represent the two most popular XAI methods based on the current literature in different domains~\cite{holzinger2022explainable}. To further substantiate this and inspired by~\cite{holzinger2022explainable}, we considered the GitHub Star, an index used for quantifying the popularity of tools on GitHub and representing appreciation and usage of tools/projects. Moreover, most of the developers consider the stars before using a specific tool~\cite{borges2018s}. Based on these considerations, we collected the GitHub Star for 10 popular XAI methods. As reported in Figure~\ref{star}, SHAP and LIME represent the most exploited methods, both featuring an increasing number of stars. Accordingly, they were considered in this perspective for the discussion of the outcomes relying on different models in two case studies.
SHAP~\cite{lundberg2017unified} is an XAI method based on game theory. It aims at explaining any model by considering each feature (or predictor) as a player and the model outcome as the payoff. SHAP provides local and global explanations, meaning that it has the ability to explain the role of the features for all instances and for a specific instance. LIME~\cite{ribeiro2016should} is another XAI method that aims at explaining how the model works locally for a specific instance in the model. To this end, it approximates any complex model and transfers it to a local interpretable model for a specific instance. Table~\ref{comparison} shows a direct comparison between both methods using different metrics. The tables shows that SHAP has some advantages over LIME. SHAP considers different combinations to calculate the features attribution while LIME fits a local surrogate model. Moreover, SHAP provides both global and local explanation while LIME is limited to local explanations only. In addition, SHAP might has the ability to detect non-linear associations (depending on the used model) while LIME fails to capturing such association because it fits a local linear model. In terms of visualization, SHAP generates several plots reporting the outcomes both locally and globally while LIME generates one plot per instance. Finally, LIME is much faster than SHAP especially with tree-based models. 

\begin{table}[H]
\centering
\caption{Comparison between SHAP and LIME.}\label{comparison}
{\begin{tabular}{|l|cc|}
\hline
\textbf{Metrics}                      & \multicolumn{1}{c|}{\textbf{SHAP}}    & \textbf{LIME}       \\ \hline
\textbf{Concept}                      & \multicolumn{1}{c|}{Applies to the model as-is}  & \begin{tabular}[c]{@{}c@{}}Fits a local surrogate model\\ to explain the complex model\end{tabular} \\ \hline
\textbf{Theory}                       & \multicolumn{1}{c|}{\begin{tabular}[c]{@{}c@{}}Additive feature attribution\\  based on game theory\end{tabular}} & Feature perturbation method                                                                                  \\ \hline
\textbf{Type}                         & \multicolumn{2}{c|}{Post-hoc model-agnostic}   \\ \hline
\textbf{Data type}                    & \multicolumn{2}{c|}{Images, tabular data and signals}        \\ \hline
\textbf{Explanation}                  & \multicolumn{1}{c|}{Global, local}  & Local    \\ \hline
\textbf{\begin{tabular}[c]{@{}l@{}} Collinearity \\ consideration\end{tabular}} & \multicolumn{1}{c|}{Not in the original method}    & No  \\ \hline
\textbf{\begin{tabular}[c]{@{}l@{}}Non-linear  \\ decision\end{tabular}} & \multicolumn{1}{c|}{Depends on the used model}   & Incapable  \\ \hline
\textbf{Computing time}               & \multicolumn{1}{c|}{Higher}  & Lower \\ \hline
\textbf{Visualization}               & \multicolumn{1}{c|}{\begin{tabular}[c]{@{}l@{}}Waterfall, Beeswarm\\and Summary plots\end{tabular}}  & One single plot \\ \hline
\end{tabular}}
\end{table}
Besides the self-explaining properties mentioned in Table~\ref{comparison}, it is worth pointing out that features collinearity and non-linear dependency across features still impact on the outcomes of both methods, limiting their reliability and, in consequence, trust. As for collinearity, even though in SHAP this issue is attenuated by the interplay of feature in and across coalitions, it still remains unsolved. In particular, the Shapley method suffers from inclusion of unrealistic data instances when features are correlated. To simulate that a feature value is missing from a coalition, it is marginalized and missing values are obtained by sampling from the feature’s marginal distribution. However, this makes sense only if features are uncorrelated \cite{learning2019guide}. In LIME the features are treated as they were independent, calling for new solutions accounting for their interplay. Along the same line, non-linear dependencies among features cannot be accounted for by LIME locally, being the local and linear surrogate model. Despite the limitations of SHAP and LIME in terms of uncertainty estimates, generalization, non-linear dependencies (with LIME), feature dependencies, and inability to infer causality~\cite{molnar2022general}, they hold substantial value for explaining and interpreting complex machine learning models.\\
However, does the end-user understand how these XAI methods work? And why they identify specific features as more informative than others? Is it enough for the end-user to know that these features are more informative, because they improve the model output without knowing how the XAI method came up with such results? For example, when SHAP assigns a high/low score for a feature, does the end user know how this score was calculated? SHAP and LIME perform many analyses in the background and solve complex equations to come up with their explanations. In many settings, complex models will be interpreted by non-expert end-users, who may find understanding of the working of XAI methods challenging. It is not expected that the end-users from different domains understand every minutiae of XAI methods, but it is vital that they are aware of the general framework of the XAI method used. While XAI methods aim at unveiling the complexity of complex black box models, they suffer from the same issue, in that their usefulness may be limited by the difficulty in understanding their outputs. In this perspective piece, we discuss SHAP and LIME XAI methods, highlighting their underlying assumptions and with the aim of helping the end-users to grasp their key concepts appropriately. We will also present some notions to increase the understanding of XAI methods and promoting their appropriate usage by the researcher community.
\subsection{SHAP}
\begin{figure}[H]
\centering
\includegraphics[width=\linewidth]{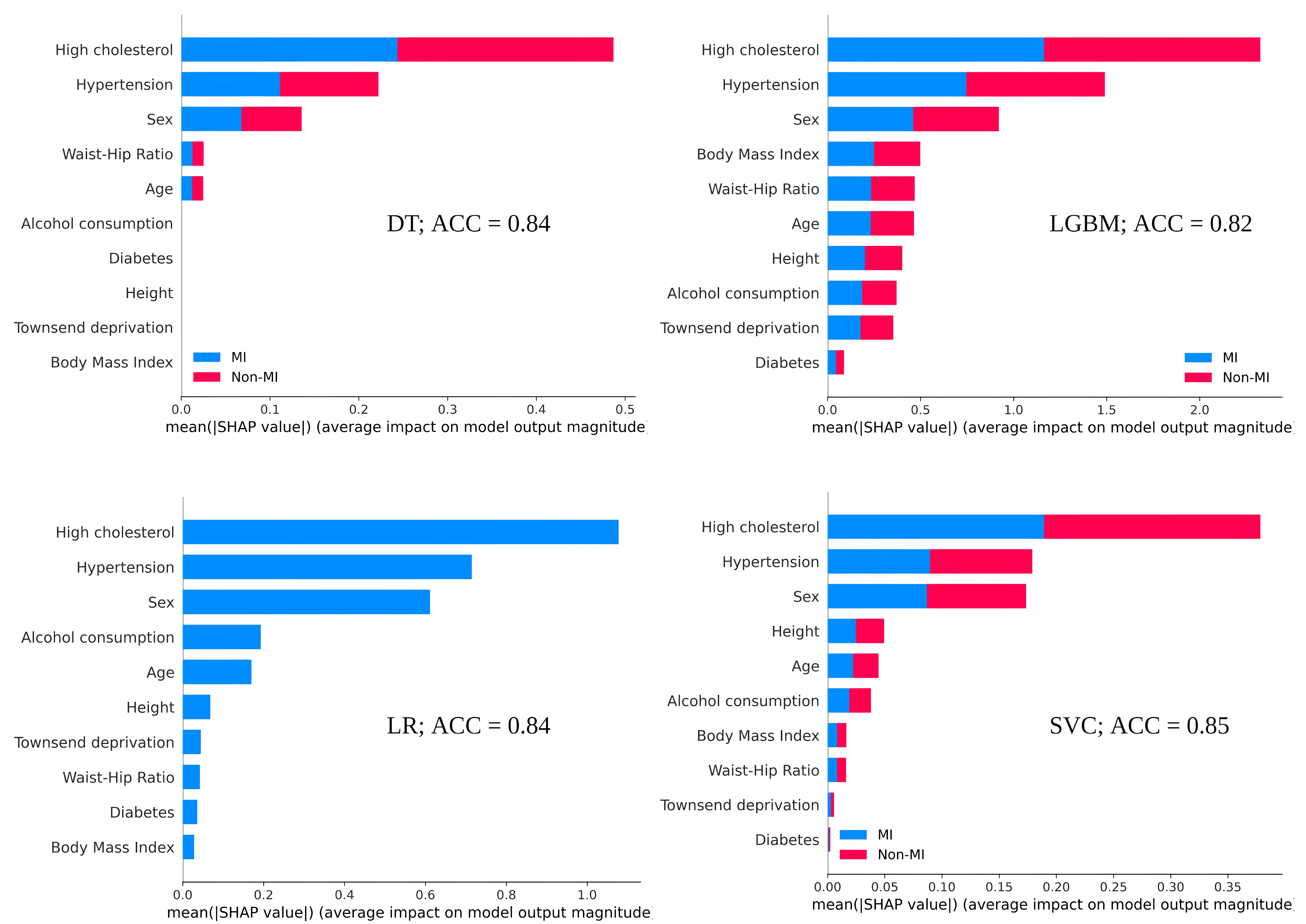}
\caption{SHAP output to explain the four models globally. DT: decision tree; LGBM: light gradient-boosting machine; LR: logistic regression; SVC: support vector machines classifier; ACC: accuracy; MI: myocardial infarction.}\label{SHAP}
\end{figure}

SHAP is a post-hoc model-agnostic method that can be applied to any machine learning model~\cite{lundberg2017unified}. It is based on game theory which calculates the contribution of each player to the payout. In machine learning models, the players and the payout are substituted by features and the model outcome, respectively. SHAP calculates a score for each feature in the model, which represents its weight to the model output. To calculate the scores, it considers all combinations between the features (i.e. coalitions) to cover all cases where all features and a subset of features are used in the model. Due to the increases of computational complexity of SHAP when the number of features increases, an approximation has been proposed, named Kernel SHAP~\cite{lundberg2017unified}.\\
SHAP has been applied widely in a variety of domains in order to explain models' outcomes, either locally or globally~\cite{garcia2020shapley, kim2022explainable, ullah2022prediction, kannangara2022investigation}. However, there are some important points the end-users should be aware of when applying SHAP. Firstly, SHAP is a model-dependent method. This means that the SHAP outcome depends on the ML model used for the classification/regression task, which will possibly lead to different explainability scores. Accordingly, when different models are applied, to the same task using the same data, the top features identified by SHAP may differ between ML models.\\
To illustrate the model-dependency point, we used four ML models to classify 1500 subjects (20\% test) from the United Kingdom Biobank into individuals with myocardial infarction (MI) and controls (Non-MI). The included models are decision tree (DT), logistic regression (LR), light gradient-boosting machine (LGBM) and support vector machines classifier (SVC). Ten different variables were considered as features in the models. These models were implemented using Python (version 3.11.4), Scikit-learn library (version 1.3.0) and the codes of SHAP on GitHub. The code of the current perspective is available at~(\url{https://github.com/amaa11/NMR}).\\
The order of the important features of the four classification models are reported in figure~\ref{SHAP}. The figure ranks the features in order of importance based on their effect toward the model outcome. It can be appreciated that there is agreement for the top three most informative features among the tested models. However there is a notable variation in the order of the remaining seven features. For instance, body mass index is the least important one in DT and LR, while it is the third in the LGBM model and the seventh in the SVC model. The position of alcohol consumption and Waist-Hip ratio similarly varies across the models. In addition, the last five features have a SHAP score close to zero in DT model, indicating they do not affect the model output. It is worth noting that despite the observed variance in feature order, the accuracy is comparable across the four classification models.

Secondly, another potential pitfall is related to the misinterpretation of the scores or SHAP values. The assigned scores do not represent the weight of the features with respect to the outcome, as their importance is encoded in the ranking. The end-users should focus on the order of the features which represents their significance. Third, SHAP is not protected against biased classifier and might generate unrealistic explanations that do not capture the underlying biases~\cite{slack2020fooling}. Finally, SHAP assumes the features are independent, thus that there is no correlation between the variables included in the ML models. In the considered case study, most of the features are collinear including high cholesterol and body mass index. Such assumption will affect the assigned score (weight) to each feature. Indeed, some features might be assigned a low score despite being significantly associated with the outcome. This is because they do not improve the model performance due to their collinearity with other features whose impact has already been accounted for. Although there are some works which tried to deal with the issue of collinearity~\cite{aas2021explaining, mase2019explaining}, yet the proposed methods are either limited to a local explanation~\cite{aas2021explaining} or the explanation is user-dependent~\cite{mase2019explaining}. Another approach was proposed to assess the stability of the list of informative features generated by XAI methods, particularly when the features are collinear~\cite{salih2022investigating}. The method calculates a value named Normalized Movement Rate (NMR) which assesses how the order of the features will be affected when the top features are removed from the model iteratively. The smaller the NMR, the more the list of informative features is stable. The authors of NMR extended their work by presenting a new method to address the collinearity issues with XAI methods. The method is named Modified Index Position (MIP). It takes the outcomes (e.g., list of informative predictors) of any XAI (e.g., SHAP, LIME) and re-order them considering the multicollinearity~\cite{salih2024characterizing}. Unlike~\cite{aas2021explaining} and \cite{mase2019explaining}, the method does not require any intervention from the user and can be applied to any model. It works similarly to NMR by iteratively removing the top feature and retrain and test the model. Thereafter, it examines how the features are re-ordered in the model which implies the effect of collinearity. More details on the method and how can be applied reflects at (\url{https://github.com/amaa11/MIP}).
\subsection{LIME}
\begin{figure}[H]
\centering
\includegraphics[width=\textwidth,height=\textheight,keepaspectratio]{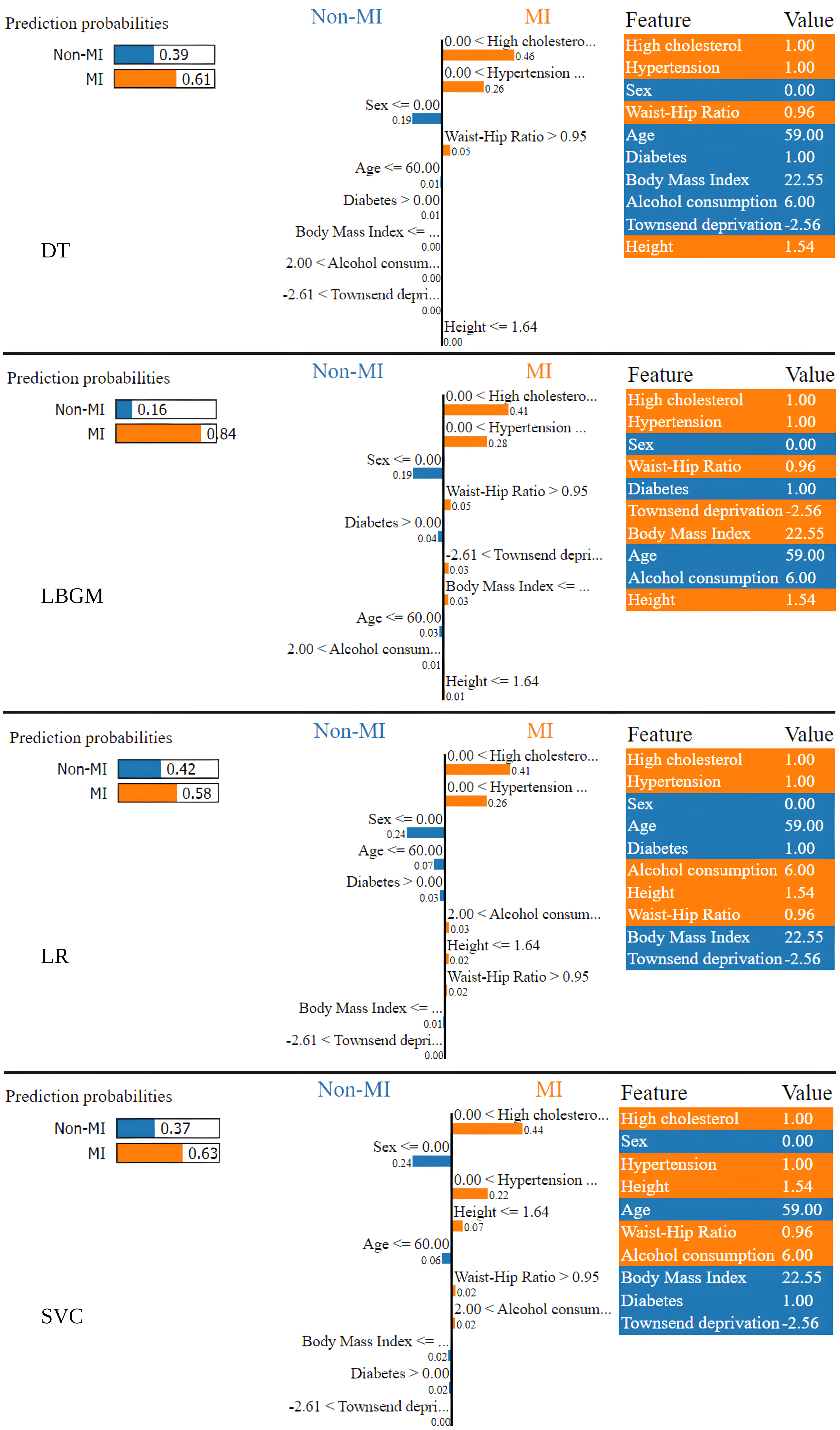}
\caption{LIME output to explain the model locally for the same instance using four classifiers. DT: decision tree; LGBM: light gradient-boosting machine; LR: logistic regression; SVC: support vector classifier; MI: myocardial infarction.}\label{LIME}
\end{figure}

LIME is a model agnostic local explanation method~\cite{ribeiro2016should}. It explains the influence of each feature to the outcome for a single subject. In the classification models, it shows the probability that the subject might belong to any class. In addition, it shows the contribution of each feature in each class with a visualized plot. \\
However, LIME converts any model into a linear local model, and then reports the coefficient values which represent the weights of the features in the model. In other word, if the user applies some models that take into account the non-linearity between features and the outcome, this might be missing in the explanation generated by LIME. This is because the non-linearity is lost in the surrogate model. In addition, LIME is a model-dependent method, meaning the used model will affect the outcome of LIME for the same task and dataset. As for SHAP, we used the same case study to evaluate the list of informative features associated with the four classifiers. Figure~\ref{LIME} shows the output of LIME for a representative subject. The first part of the plot (left) shows the probability that the subject is classified as control (Non-MI) or with MI in each of the used models. The second part (middle) shows the weight, i.e. coefficient value, of each feature in the local linear model,  while the last part on the right shows the actual value of each feature. Moreover, the plot shows the features contribution toward each class based on the assigned color. In this case, the probability belong to one or the other class is different for each of the used models. It shows that LGBM is the most certain, while DT is the least. In addition, the plot shows that the same feature is contributing to different classes across the tested models. For example, alcohol consumption contributes to the MI class in LR and SVC, while it contributes to the Non-MI class in the DT and LGBM. 

Body mass index and Townsend deprivation contribute to the MI class in the LGBM model, while they contribute to the Non-MI class in the other three models. In addition, the used features have similar effect size although four different models were used. This is due to the fact that LIME generates and approximate a local linear model and then reports the weights of the features. 
Concerning collinearity, the interpretation of the weights generated by LIME indicates that an increase/ decrease per one unit change in the feature will lead to an increase/decrease in the outcome while other features are kept unchanged. Such assumption is not realistic with collinear data where groups of features might change simultaneously. It is indeed the correct interpretation for the coefficient values in linear models. But because they are generated by LIME, the user might think that they have more power and meaning than the classical coefficient values in the machine learning models. Finally, similarly to SHAP, LIME can be fooled by biased classifiers, leading to explanations that do not reflect or represent the biases~\cite{slack2020fooling}.
\section{A case study}
\begin{figure}[H]
\centering
\includegraphics[width=\textwidth]{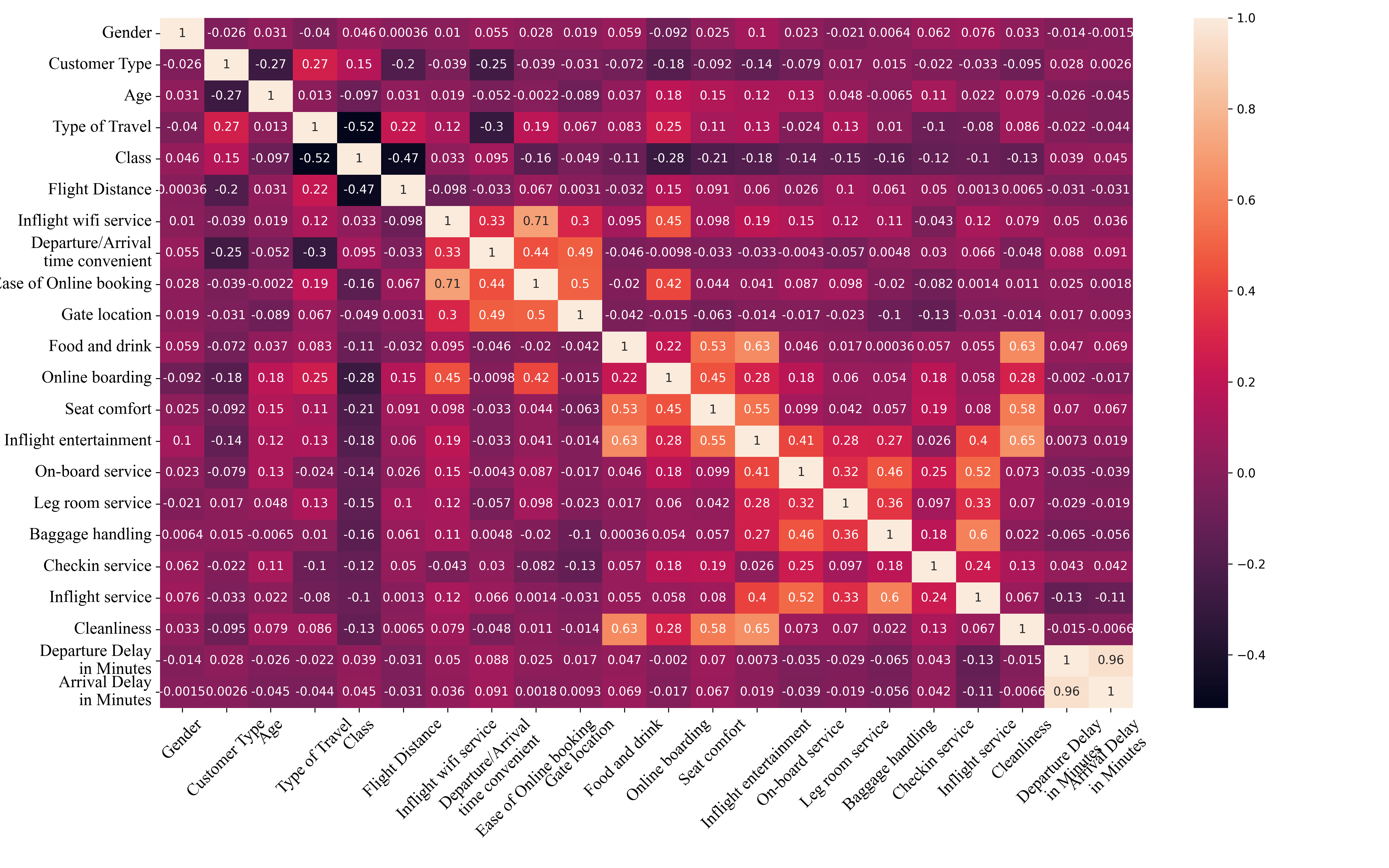}
\caption{Correlation heatmap.}\label{correlation}
\end{figure}

The following case study illustrates the limitations of SHAP in terms of model dependency and collinearity, and the possible available solutions to overcome them. The case study can be extended to LIME as well as to any other XAI method. \href{https://www.kaggle.com/datasets/teejmahal20/airline-passenger-satisfaction}{Airline Passenger Satisfaction} data from Kaggle for 500 subjects (satisfied, n=250) was used in the case study. Out of these, 22 features were used to predict whether the passenger was satisfied or not. Four classifications models were used that are LGBM, LR, DT and SVC. The data was divided into training and testing (20\%). Default parameters of the each model were used. Thereafter, SHAP was applied to identity the most informative predictors for each model. Figure~\ref{correlation} shows the correlation heatmap of the used features in the model. The figures shows that there is collinearity between some features which will affect the outcome of XAI.
Table~\ref{SHAPcase} shows the most informative features in each model generated by SHAP. It is noted that each model generated a different list of informative features although their accuracy were relatively similar apart from LGBM for which it is higher. It is worth to mention that we cannot be certain the LGBM is better than the other models because we used the default parameters of each model, and applying hyperparameter tuning might produce different accuracy for each model. The variation in the list even observed in the top one where two models identified \textit{Class} as the most important one while the other two identified other features. The question is which one of these list to consider given that the data are collinear and each model presented a different list generated by SHAP. Some might argue that we should consider the outcome of SHAP with LGBM as LGBM reached the highest accuracy among other models. However, in the previous case we showed that the accuracy of the models might be comparable with some data.
\begin{table}[ht]
\centering
\begin{tabular}{|c|c|c|c|}
\hline
\textbf{LGBM (ACC: 0.91)}&                 \textbf{LR (AC: 0.85)}                & \textbf{DT (ACC: 0.84)}            & \textbf{SVC (ACC: 0.86)}                                  \\ \hline
Inflight wifi service      & Class         & Online boarding           & Class                                                          \\ \hline
Type of Travel   & Online boarding    & Inflight wifi service & Online boarding                                                          \\ \hline
Online boarding    & Cleanliness  & Type of Travel  & Type of Travel                                                                     \\ \hline
Class & Seat comfort    & Class   & Seat comfort                                                                                         \\ \hline
Cleanliness    & Arrival Delay in Minutes   & Cleanliness      & Cleanliness                                                             \\ \hline
On-board service  & Inflight wifi service      & Age      & Inflight wifi service                                                        \\ \hline
\begin{tabular}[c]{@{}c@{}}Departure/\\ Arrival time convenient\end{tabular} & Customer Type   & Leg room service   & Leg room service   \\ \hline
Baggage handling & \begin{tabular}[c]{@{}c@{}}Departure Delay\\  in Minutes\end{tabular}  & Customer Type  & Food and drink   \\ \hline
Leg room service  & Ease of Online booking   & Inflight entertainment   & On-board service       \\ \hline
Food and drink & Type of Travel         & Baggage handling  & Arrival Delay in Minutes                                                   \\ \hline
Age    & Gender     & Gender           & Ease of Online booking                                                \\ \hline
Customer Type                                  & On-board service                                                             & On-board service                                                             & Departure Delay in Minutes                                                   \\ \hline
Flight Distance                                                              & Flight Distance                                                              & Arrival Delay in Minutes                                                     & Gate location                                                                \\ \hline
\begin{tabular}[c]{@{}c@{}}Arrival Delay\\  in Minutes\end{tabular}          & Leg room service                                                             & Gate location                                                                & Gender                                                                       \\ \hline
Ease of Online booking                                                       & Age                                                                          & Flight Distance                                                              & Inflight service                                                             \\ \hline
Seat comfort                                                                 & \begin{tabular}[c]{@{}c@{}}Departure/\\ Arrival time convenient\end{tabular} & Seat comfort                                                                 & Baggage handling                                                             \\ \hline
Gate location                                                                & Inflight entertainment                                                       & \begin{tabular}[c]{@{}c@{}}Departure Delay\\ in Minutes\end{tabular}         & Checkin service                                                              \\ \hline
Inflight service                                                             & Food and drink                                                               & Ease of Online booking                                                       & Customer Type                                                                \\ \hline
Checkin service                                                              & Inflight service                                                             & Inflight service                                                             & Flight Distance                                                              \\ \hline
\begin{tabular}[c]{@{}c@{}}Departure Delay\\ in Minutes\end{tabular}         & Gate location                                                                & Food and drink                                                               & \begin{tabular}[c]{@{}c@{}}Departure/\\ Arrival time convenient\end{tabular} \\ \hline
Gender                                                                       & Checkin service                                                              & Checkin service                                                              & Inflight entertainment                                                       \\ \hline
Inflight entertainment                                                       & Baggage handling                                                             & \begin{tabular}[c]{@{}c@{}}Departure/\\ Arrival time convenient\end{tabular} & Age                                                                          \\ \hline
\end{tabular}
\caption{List of informative features produced by SHAP. LGBM: light gradient-boosting machine; logistic regression; DT: decision tree; LR: SVC: support vector machines classifier; ACC: accuracy.}
\label{SHAPcase}
\end{table}
NMR helps to examine which one of these models produced a more stable list against the collinearity. We have applied NMR to each model which produced the following results: LGBM: 0.231, LR: 0.275, DT: 0.445 and SVC: 0.273. Accordingly, LGBM has the lowest NMR value which indicates that the corresponding outcome is the most robust. NMR shows which model is more stable but it does not enhance the outcomes of SHAP to consider the collinearity. MIP then can be used to modify the outcome of SHAP and to obtain a list of informative features that consider the dependency among the features. The outcome of MIP for LGBM (with smallest NMR value) alongside with the SHAP outcome is explained in Table~\ref{SHAP_MIP}. The table shows that there is variation in each list. For example, \textit{Ease of Online booking} is the fifteenth in SHAP list while it is the fifth when MIP was applied.

\begin{table}[ht]
\centering
\begin{tabular}{|c|c|}
\hline
\textbf{SHAP}                     & \textbf{MIP}                      \\ \hline
Inflight wifi service             & Inflight wifi service             \\ \hline
Type of Travel                    & Online boarding                   \\ \hline
Online boarding                   & Type of Travel                    \\ \hline
Class                             & Class                             \\ \hline
Cleanliness                       & Ease of Online booking            \\ \hline
On-board service                  & On-board service                  \\ \hline
Departure/Arrival time convenient & Cleanliness                       \\ \hline
Baggage handling                  & Arrival Delay in Minutes          \\ \hline
Leg room service                  & Inflight entertainment            \\ \hline
Food and drink                    & Leg room service                  \\ \hline
Age                               & Food and drink                    \\ \hline
Customer Type                     & Flight Distance                   \\ \hline
Flight Distance                   & Departure/Arrival time convenient \\ \hline
Arrival Delay in Minutes          & Seat comfort                      \\ \hline
Ease of Online booking            & Baggage handling                  \\ \hline
Seat comfort                      & Age                               \\ \hline
Gate location                     & Departure Delay in Minutes        \\ \hline
Inflight service                  & Inflight service                  \\ \hline
Checkin service                   & Gate location                     \\ \hline
Departure Delay in Minutes        & Checkin service                   \\ \hline
Gender                            & Customer Type                     \\ \hline
Inflight entertainment            & Gender                            \\ \hline
\end{tabular}
\caption{List of informative features produced by SHAP and modified by MIP}
\label{SHAP_MIP}
\end{table}
We have applied MIP to produce a global list of informative features. In a similar way, if the aim is to provide a local list of informative features, then XAI should be applied locally for a specific subject followed by MIP. In addition, local explanation for a specific individual can also be produced by applying the proposed method~\cite{aas2021explaining} that modifies SHAP to consider the collinearity.
\section{Recommendations}
SHAP and LIME are two popular XAI methods that aid understanding ML models in different research fields. They have been implemented in some sensitive domains~\cite{george2021deep, haimovich2020development, zhang2022applications} where misinterpreting the outcomes might be very expensive or critical. Data scientists who are working daily on ML and XAI tend to over-trust the explanations generated by XAI methods and do not accurately understand the visualized output of the XAI methods~\cite{kaur2020interpreting}, that could result in a misuse of the interpretability tools.\\
It is crucial that SHAP results are presented alongside the corresponding output plots, presenting them with a simple language to explain the outcomes and the assumptions behind SHAP (e.g. features are independent and the outcomes are model-dependent). Moreover, if possible, the end-users should implement different ML models when dealing with collinear features in order to compare the SHAP outcomes across models and evaluate their robustness. Using post-hoc proxies such as the NMR~\cite{salih2022investigating} value would be useful to select the model that presents the more stable list of informative features generated by any XAI method. MIP~\cite{salih2024characterizing} then can be used to enhance the outcome of XAI in presence of collinear features if the aim is to explain the model globally. On the other hand, if the aim is to explain the model locally for a single instance or sub-group of individuals, then MIP~\cite{salih2024characterizing} and approximated SHAP value (shapr)~\cite{aas2021explaining} can be implemented. This is because MIP can be applied to any XAI method and shapr is a modified version of SHAP, and both takes into account the collinearity among the features. In addition, converting the scores of SHAP of each feature of the model (especially in classification models) into a more digestible form would increase the understanding of the score and ultimately the method itself. It is worthy to note that LIME provides explanation regarding the local model linearity with the model outcome as the users might not be familiar with the concept behind LIME. The users will be more aware and understand the outcome when a simple language accompanies the outcome. Moreover, the explanation of LIME might be different using the same model, but for other instance. In other words, the interpretation of LIME only applies for one subject and cannot be used or considered as a general interpretation for the whole model. Finally, GraphLIME~\cite{huang2022graphlime} was proposed as an updated version of LIME to explain graph-based models where non-linear association is more appropriately considered.
\section{Conclusions}
In the current perspective, we discussed two widely used XAI methods specially with tabular data. The highlighted and discussed points are very significant and critical to be considered when XAI methods are implemented in any domain. Considering the end-users not from technical background, it is needful that they are aware of these issues in order to use the methods most appropriately.

\section{Funding}
AMS is supported by a British Heart Foundation project grant (PG/21/10619). IBG and GM acknowledge support from Fondazione CariVerona (Bando Ricerca Scientifica di Eccellenza 2018, EDIPO project - reference number 2018.0855.2019. ZRE recognises the National Institute for Health and Care Research (NIHR) Integrated Academic Training programme which supports her Academic Clinical Lectureship post and was also supported by British Heart Foundation Clinical Research Training Fellowship No. FS/ 17/81/33318. SEP acknowledges support from the National Institute for Health and Care Research (NIHR) Biomedical Research Centre at Barts and have received funding from the European Union’s Horizon 2020 research and innovation programme under grant agreement No 825903 (euCanSHare project). This article is supported by the London Medical Imaging and Artificial Intelligence Centre for Value Based Healthcare (AI4VBH), which is funded from the Data to Early Diagnosis and Precision Medicine strand of the government’s Industrial Strategy Challenge Fund, managed and delivered by Innovate UK on behalf of UK Research and Innovation (UKRI). Views expressed are those of the authors and not necessarily those of the AI4VBH Consortium members, the NHS, Innovate UK, or UKRI.

\end{document}